\documentclass[preprint,12pt]{elsarticle}

\usepackage{amssymb}
\usepackage{amsmath}
\usepackage[utf8]{inputenc} 
\usepackage[T1]{fontenc}    
\usepackage{hyperref}       
\usepackage{url}            
\usepackage{booktabs}       
\usepackage{amsfonts}       
\usepackage{nicefrac}       
\usepackage{microtype}      
\usepackage{lipsum}		
\usepackage{graphicx}
\usepackage{natbib}
\usepackage{doi}
\usepackage{graphicx}
\usepackage{float}
\usepackage{subcaption}
\usepackage{pgfplots}
\usepackage[strings]{underscore}

\journal{Computer Speech \& Language}

\begin{document}

\begin{frontmatter}

\title{Hybrid Dialogue State Tracking for Persian Chatbots: A Language Model-Based Approach}


\author{Samin Mahdipour Aghabagher, Saeedeh Momtazi \\ Computer Engineering Department, Amirkabir University of Technology}
\ead{momtazi@aut.ac.ir}
%
%

\begin{abstract}
 Dialogue State Tracking (DST) is an essential element of conversational AI with the objective of deeply understanding the conversation context and leading it toward answering user requests. Due to high demands for open-domain and multi-turn chatbots, the traditional rule-based DST is not efficient enough, since it cannot provide the required adaptability and coherence for human-like experiences in complex conversations. This study proposes a hybrid DST model that utilizes rule-based methods along with language models, including BERT for slot filling and intent detection, XGBoost for intent validation, GPT for DST, and online agents for real-time answer generation. This model is uniquely designed to be evaluated on a comprehensive Persian multi-turn dialogue dataset and demonstrated significantly improved accuracy and coherence over existing methods in Persian-based chatbots. The results demonstrate how effectively a~hybrid approach~may improve DST capabilities, paving the way for conversational AI systems that are more customized, adaptable, and human-like.
\end{abstract}





\end{frontmatter}









\section{Introduction}
Chatbots have become an inseparable part of our daily lives, helping with everything from personal tasks to industrial customer service. With conversational AI as assistants, it is possible to offer 24/7, fast, and accurate support with fewer service fees and no human resource costs. Thus, many companies are investing in these technologies, as their user-satisfactory and cost-effective benefits contribute to a competitive edge in the market. In the beginning, chatbots were designed to automatically answer users' pre-considered questions, but today their abilities cover a wide range of duties that do not demand physical requirements, including making medical appointments, handling post-related tasks, recommending products based on the user's needs and preferences, responding to complaints, etc \cite{Lee2020}. However, as people's demands are increasing, it is expected that chatbots will become something more than just a simple question-answering system that was resolved many years ago. Chatbots are challenged by many factors, such as customization, multilingualism, humanization, and many other critical issues.

Dialogue State Tracking (DST) is an essential component of conversational AI systems that determines the current state of the dialogue flow. The dialogue state is not just the slot fillers in the current sentence; it includes the entire state of the frame at this point, summarizing all of the user's constraints \cite{Jurafsky2008}. In other words, by tracking the states of the conversation and managing the slot-value pairs, DST leads the conversation toward the objective of responding. DST is also key to making chatbots more human-like, as human conversations naturally maintain their flow and aim to achieve their goals as efficiently as possible.

Despite other main modules, such as intent detection and slot filling, which are focused on each turn individually, DST considers the dialogue in both singular and overall aspects. Firstly, it handles all the user's information throughout the conversation as slot-value pairs, and secondly, based on the goal of the conversation, DST leads the conversation by asking an appropriate question to gather the required information to propose an optimized response as soon as possible. As a result, DST creates a more humanized experience and provides the most contextually relevant answers that enhance the user's satisfaction.

Chatbots can be divided into two groups: open-domain dialogue systems, which handle general conversational topics, and task-oriented dialogue systems, designed to assist users with specific tasks \cite{Niu2024}. A human-based conversation is not bound by any rules or topics so task-oriented dialogue systems cannot be effective any longer if one wants to improve chatbot technologies to be more customized and human-like. On the other hand, this is a great challenge, as it requires comprehensive datasets and highly advanced DST models that can track the conversation without any particular rule. 

Traditional DST methods utilized rule-based solutions, which were only useful in the limited-domain chatbots. As mentioned, these systems lack the required adaptability for a complex, human-like conversation that can be seen in real-world applications. Thus, designing a dynamic, consistent module that can handle open-domain conversations remains a research challenge.

On the other hands, not all the people in the world speak the same language. In the basic conversational systems, which were not designed to be used for all members of the community, this was not a real challenge, but now the main objective of AI assistants is to be as customized as required to help all the people in any possible way. Therefore, it is vital to train conversational AI models that can handle multilingualism. This requires an enormous dialogue-based dataset for each language and great computational resources to train models on them. Moreover, each language has its grammar, alphabet, and complexity. If a conversational AI assistant is designed to be useful, it must gain expertise in that specific language. 

This study proposes a hybrid approach for DST in Persian chatbots that leverages language models (such as BERT and GPT) along with rule-based methods. It tries to combine the precision and interpretability of rule-based approaches with the adaptability and generative powers of  Large Language Models (LLMs), hence improving chatbots' flexibility and consistency in open-domain, multi-turn interactions. For this purpose, a comprehensive Persian dialogue-based dataset is also generated with the structure of the Wizard-of-Oz dataset, which is completely unique among Persian datasets.
    
The rest of the paper is organized as follows: Section~\ref{sec:relatedworks} presents the related works on Dialogue State Tracking challenges, proposed models, and dialogue-based datasets. Section~\ref{sec:dataset} describes the Persian datasets that were developed for this study. Section~\ref{sec:proposedmodel} details the proposed hybrid model, including its modules such as NLU, intent validation, dialogue state tracking, and online agents. It also discusses the challenges that this model addresses. Section~\ref{sec:evaluation} reports the experimental setup and evaluation results. Finally, Section~\ref{sec:conclusion} concludes the paper and discusses directions for future research.

\section{Related Works}
\label{sec:relatedworks}

To train LLMs for DST, it is essential to access comprehensive dialogue-based datasets. 

\textbf{The Wizard of Oz (WOZ) Dataset} is one of the most comprehensive and widely used datasets in the field of conversational systems. It is designed to train and evaluate natural language processing models and conversational systems, and due to its diversity and comprehension, it is widely used by researchers and developers \cite{Budzianowski2018}. This dataset is named the Wizard of Oz Dataset because it acts similarly to the story, \textit{The Wonderful Wizard of Oz} by \citet{baum1900oz}. In this story, a human acts as the system behind the scenes, and the user imagines that he is interacting with a real computer system.

In \textbf{PerSHOP - A Persian Dataset for Shopping Dialogue Systems Modeling}, \citet{Mahmoudi2024} introduced a Persian dataset for shopping dialogue systems modeling, which includes conversations and sentences in various domains, such as clothing, vehicles, food, etc. The data is collected with the participation of users to ensure the diversity and accuracy of the conversations. Modeled on the Wizard of Oz dataset, this dataset provides structured and labeled conversations with specific intentions and gaps that can be used to train conversational shopping systems models.

DST is crucial for conversational agents as it facilitates the comprehension and management of context and intent in human interactions. \citet{Henderson2015} in \textbf{Machine Learning for Dialog State Tracking: A Review} identified the main challenge of DST as the need to sustain an accurate depiction of dialogue states across various phases, a task made difficult by ambiguous user input and the fluid nature of conversations. The simplicity and interpretability of traditional techniques, such as rule-based and generative methods, have been demonstrated. However, the efficacy of these methods in real-life scenarios is often restricted by their inability to adapt to a variety of conversation contexts and their lack of scalability. This research delineates a variety of machine learning algorithms that have been proposed to address these challenges, illustrating significant advancements over traditional methods. Static classifiers, which are dependent on immutable feature sets, frequently underperform in dynamic scenarios. Conversely, sequence models, such as recurrent neural networks, enhance the ability to manage temporal context. By integrating the results of numerous models, system combination methods improve the efficacy of decision support systems. Additionally, methodologies for modeling joint distributions enhance state predictions by integrating the interrelationships among discourse components. Unsupervised adaptation and generalization techniques show the potential to decrease dependence on labeled data, thereby expanding the applicability of DST systems across multiple domains. The evolution of DST capabilities is evidenced by the shift from rule-based approaches to sophisticated machine learning techniques, facilitating the incorporation of LLMs in forthcoming research endeavors.

In a subsequent work titled \textbf{Deep Neural Network Approach for the Dialog State Tracking Challenge}, \citet{Henderson2013} proposed a deep neural network (DNN) model for Dialogue State Tracking (DST). In previous papers, generative methods were used to determine the state of a conversation, but this research made a pivotal shift to discriminative learning. Motivated by recent advancements in deep learning within speech research, the paper proposes a model that utilizes deep neural networks (DNNs) to capture complex interactions among dialog features, leading to improved performance over traditional approaches with fewer hidden layers. The model outputs a sequence of probability distributions over possible values using a single neural network, allowing for direct maximization of the log-likelihood of the training data through gradient ascent techniques. This DNN model indicated a significant performance improvement compared to previous models. Its advantages include the remarkable ability to learn the conversation's features, robust generalization, and reduced dependency on training data, leading to effective performance even under previously unseen or noisy conditions.

\citet{Feng2023} investigated the potential of LLMs, such as ChatGPT, and smaller, open-source alternatives to be employed in DST in \textbf{Towards LLM-driven Dialogue State Tracking}. The authors discuss the increasing interest in LLMs, particularly due to their capability to manage complex language patterns in multi-participant conversations. Although ChatGPT demonstrates good performance, it poses considerable challenges, such as being a closed source, enforcing usage restrictions, and generating privacy concerns related to data management. LLM-driven DST framework (LDST), a novel system that has been fine-tuned for DST and utilizes smaller, open-source foundation models, was developed by \citet{Feng2023} in response to these challenges. It has demonstrated effective performance in both zero-shot and few-shot scenarios.  The framework employs domain-slot instruction tuning. This method enhances DST performance by instructing the model to comprehend domain-specific instructions. LDST outperformed the previously established best methods. Significant advancements were observed in both zero-shot and few-shot learning tasks. This paper demonstrates that LLM-based methods can enhance the accuracy of DST and simultaneously tackle practical challenges associated with the use of closed-source models such as ChatGPT.

\citet{Heck2023} proposed ChatGPT as a general-purpose model that may handle zero-shot DST tasks without fine-tuning. \textbf{ChatGPT for Zero-shot Dialogue State Tracking: A Solution or an Opportunity?} examines ChatGPT's efficacy as a DST model in task-oriented talks, given its wide dataset training and lack of domain-specific training. ChatGPT has 31.5\% join goal accuracy (JGA) in zero-shot settings on MultiWOZ 2.1. This performance is impressive for a machine not trained for DST. According to the authors, ChatGPT is particularly adept at managing co-references and extracting slot values directly from user input, particularly in complex dialogues with numerous exchanges. The model is capable of learning from its surroundings, which enables it to perform effectively in areas that have not been previously encountered. In these regions, other supervised models may face challenges as a result of their lack of exposure to specific slot values. ChatGPT outperforms other models in "don't care" slots and co-reference resolution for underrepresented groups. The study shows that general-purpose models have limitations despite promising results. ChatGPT may generate erroneous data or fake normalizations, reducing its practical reliability. It struggles to transfer system-mandated data and resolve partial references. Specialized DST models demonstrate a higher level of reliability in this regard. ChatGPT is effective at dealing with singular problems, but it cannot entirely replace domain-specific decision support systems due to its unpredictable behavior and limited precision in result control, according to the authors. By generating training data, addressing inputs outside the familiar domain, or providing real-time assistance with previously uncounted slot values, it may serve as an additional resource for specialized systems. This study recommends combining ChatGPT's generalization with classical models, which are better for certain tasks. This combination maximizes benefits: ChatGPT's adaptability to manage zero-shot operations and specialized models' precision and reliability .

\citet{Hu2022} introduced an innovative approach for DST via in-context learning (ICL) in short-shot contexts. The authors introduce a system designated as IC-DST. A test case and other instances are utilized to train an LLM in forecasting the progression of a discussion. In contrast to other methods, IC-DST functions by presenting the language model with a sequence of real-world scenarios rather than performing fine-tuning. As a result, it is highly adaptable and capable of operating in new domains without the need for retraining. This method addresses the issues of scalability and adaptability in DST, where obtaining labeled data is sometimes challenging or entails significant costs for integration. The authors explore three notable innovative characteristics of the framework. Initially, they reframe DST as a text-to-SQL assignment. This challenge illustrates the current state of the dialogue between SQL queries and databases. This format allows the system to employ pre-trained models, such as Codex, therefore expediting DST tasks. The complete history of conversations need not be incorporated within the framework. The discourse context is illustrated by state changes, which are the alterations in slot values between successive turns. This approach is more effective. Ultimately, choosing situations from real life based on the resemblance of their dialogue state transitions improves the model's precision.  The application of few-shot learning on the MultiWOZ dataset markedly enhanced the outcomes achieved with the IC-DST system. It demonstrated an efficacy increase of 10{\textendash}30 percent in attaining the collective objective vs. alternative approaches. It operates efficiently in zero-shot scenarios, necessitating no task-specific data. This research signifies notable progress for DST, illustrating that LLMs can proficiently handle intricate, multi-turn dialogues in few-shot contexts without considerable retraining.

Context is illustrated by state changes, which are the alterations in slot values between successive turns. This approach is more effective. Ultimately, choosing situations from real life based on the resemblance of their dialogue state transitions improves the model's precision. The application of few-shot learning on the MultiWOZ dataset markedly enhanced the outcomes achieved with the IC-DST system. It demonstrated an efficacy increase of 10{\textendash}30 percent in attaining the collective objective vs. alternative approaches. It operates efficiently in zero-shot scenarios, necessitating no task-specific data. This research signifies notable progress for DST, illustrating that LLMs can proficiently handle intricate, multi-turn dialogues in few-shot contexts without considerable retraining.

Even though several studies have addressed DST using deep learning and LLMs, most have focused on English datasets and often did not consider real-world dialogue challenges such as ambiguous intents, flexible answers like ``whatever,'' and dialogue topic shifts. In comparison with these models, our model introduces an intent validation module, supports Persian, and is evaluated on a custom-designed multi-turn dataset reflecting natural conversational patterns. Notably, the proposed model directly handles nearly all of the limitations and challenges outlined in these prior works, offering practical solutions for ambiguity detection, slot-value uncertainty, intent clarification, and multilingual support. This combination results in more accurate, coherent, and human-like dialogue management in open-domain scenarios. The proposed model was developed considering all challenges which was addressed in the related studies.

\section{Dataset}
\label{sec:dataset}

Existing Persian datasets were limited and could not be used for training the proposed model; therefore, it was necessary first to collect appropriate data for designing the conversational AI system and its key modules.




\subsubsection{Ontology}

In this study, we defined 20 domains, covering topics such as entertainment, economics, and etc. Every domain encompasses several intents. A total of !!! intents were established in this study. Excluding improper or out-of-domain intentions, the remaining intents can be further categorized based on the existence of slots. For slot-based intents, from one to four mandatory or optional slots were specified.





\subsubsection{Dialogue-Based Dataset}

In chatbot development, the goal goes beyond the basic detection of a single sentence's intent; it also encompasses the replication of actual human conversational flow as accurately as possible. Most Persian datasets are collected in a single-turn format. In other words, there is a collection of sentences for each intent. However, conversational AI systems require dialogue-based datasets to comprehend several aspects and the complexity of human conversations. This kind of dataset is essential for training the Natural Language Understanding (NLU) module, which is a fine-tuned BERT model in our case, to fulfill the objective of intent detection and slot filling. Furthermore, the dataset must be comprehensive to support open-domain ability. This section's dataset was developed and gathered utilizing the MultiWOZ (Multi-Domain Wizard-of-Oz) dataset as a reference architecture. The data-collecting approach for both categories corresponded to an approach similar to that employed in the construction of the MultiWOZ dataset, ensuring a realistic and cohesive conversational framework appropriate for training dialogue systems.

\subsubsection{Question Set for DST}The DST component determines the current dialogue state, including the status of mandatory slots required for answer generation. In order to gather the missing data, the system must ask follow-up questions if the user's input does not contain certain required slots.

To address this need, five question samples were created for each mandatory slot in the ontology. These inquiries are crafted to be contextually relevant and may be randomly chosen by the state tracker whenever user clarification is required throughout the dialogue.



\section{Proposed Model}
\label{sec:proposedmodel}

\subsection{Overall Structure of the Proposed Model}

The proposed hybrid architecture is structured as a multi-stage pipeline, integrating four main components: (1) a BERT-based Natural Language Understanding (NLU) module for slot filling and intent detection, (2) an XGBoost-based intent validation layer to refine and disambiguate user intents, (3) a GPT-based Dialogue State Tracker (DST) guided by optimized prompts to generate structured dialogue states and database queries, and (4) online GPT-based agents for real-time information retrieval and natural response generation. Each component plays a specialized role in ensuring that user inputs are accurately interpreted, validated, and responded to within the flow of a coherent multi-turn dialogue.

\begin{figure}[H]
    \centering
    \includegraphics[width=\textwidth]{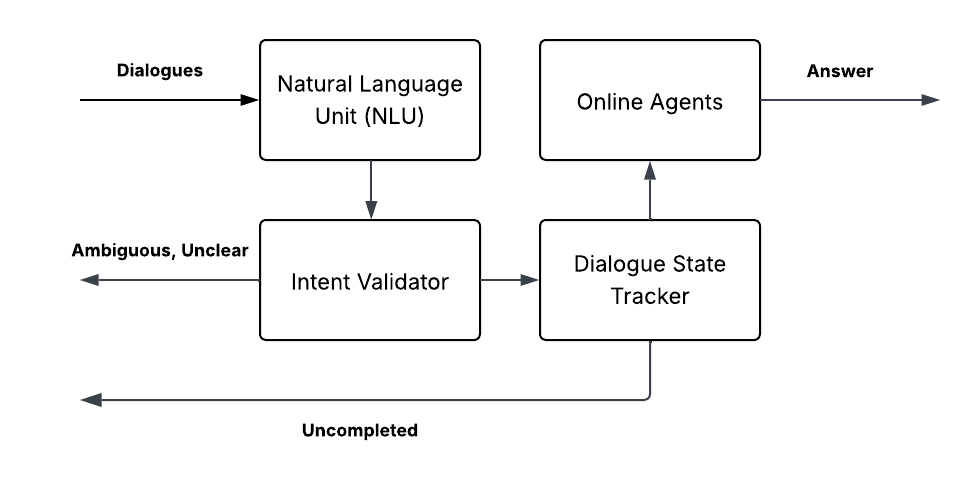}
    \caption{Overall architecture of the proposed Dialogue State Tracking model.}
    \label{fig:hybrid-architecture}
\end{figure}

\subsection{Slot Filling and Intent Detection}The initial phase of the chatbot utilizes BERT, a transformer-based language model, to extract slot values and detect the user's intent. BERT is fine-tuned with domain-specific Persian data to~ ensure accurate detection of essential slot values, such~ as locations and timings, and defined intents. By leveraging BERT's context-aware language comprehension abilities, the model establishes an excellent basis for the accurate interpretation of user input and detecting fundamental intentions, which are vital to the subsequent management of dialogue. Based on this component's outputs, other modules, such as DST and the intent validator, take action.~ BERT was chosen for its shown efficacy in several linguistic tasks, such as intent classification and slot filling.  Its capacity to model comprehensive contextual representations enables it to clarify meaning depending on surrounding tokens, which is particularly advantageous in conversation systems where intent and slot values tend to be implicit or vaguely expressed.

 Furthermore, Persian language processing presents distinct obstacles due to its complex grammar, dynamic word order, and utilization of complicated verb formations.  BERT, particularly when fine-tuned on language-specific or domain-specific Persian datasets, exhibits robust generalization about these complexities.  BERT's rich contextual embeddings enable it to capture the many subtleties and syntactic interactions that accompany Persian, in contrast to simpler models that depend significantly on handcrafted features and extensive feature engineering.

\subsection{Intent Validation}

There are many cases where NLU's recognition must be assessed. Sometimes users may mention more than one intent in their request. However, the NLU may choose one intent with a high score even though other intents also got high scores. However, users may not clearly mention their intention in their request, so NLU cannot make a reliable choice. In this study intent validator is proposed to address these kinds of challenges. 

XGBoost is utilized as a classifier to validate the recognized intents from NLU with the help of its scores. Due to a significant class imbalance in the collected dataset, detailed in the experiments and results section, XGBoost was selected for its well-established performance on imbalanced data. In addition, the model incorporates feature engineering techniques to derive informative characteristics from the input. The classifier categorizes intents into three groups: validated (if the primary intent aligns with the intent identified by BERT), ambiguous (if two intents demonstrate high scores), and unclear (if no intent fulfills the minimum score threshold). By this classification, the flow of conversation will be leading to DST if the recognized state is validated. Otherwise, the chatbot asks the user to clarify their intention better and then takes further steps.

\subsection{GPT Instruction with Optimized Prompts for DST} 

Using the ICL approach, an optimal prompt is designed for each of the intents recognized in the ontology. Using rule-based methodologies, the optimal prompt is selected for a specific intent, and GPT is instructed with this prompt to generate a JSON structure that contains the dialogue state, an SQL query for mandatory slots related to database operations, and any inquiries required for leading the conversation.

GPT was selected for this challenge due to its adaptability, generative features, and robust performance in the Persian language. Among existing LLMs, GPT demonstrates superior generating capabilities for Persian, adeptly managing complex sentence structures and vocabulary with remarkable precision. Its multilingual capabilities provide a significant advantage, enabling it to effortlessly comprehend and respond to many languages and mixed-language inputs, an aspect that is especially important for practical applications.

Another significant advantage is that GPT does not require any further adjusting to comprehend and generate text in structured forms like database queries. It follows the instructions carefully, facilitating seamless integration with other system components and database processes. The accuracy and adaptability of GPT make it an excellent base for the DST module inside the proposed model.

\subsection{Online GPT-based Agents for Answer Generation}
In this part, online agents are implemented to manage the user's request derived from the final result of DST, which is a JSON structure including the dialogue status, validated intent, and slot values, and find the required information from the web to generate an appropriate answer. GPT is used as the base LLM for this purpose, as it has great access to web-based data, and it also has the ability to generate responses using the retrieved data.

One of the advantages of GPT is the variety of tools and integrations. Frameworks like LangChain, which are developed to enable developers to construct organised prompt sequences and memory-capable systems with~ no overhead. These tools provide modular, scalable structures and allow for fast prototyping and deployment of LLM-driven systems.

One of the primary advantages of utilizing GPT is its facilitation of the development of smart online agents. These agents can independently perform actions, such as retrieving information. Utilizing libraries such as LangChain enables a more straightforward deployment. Rather than only responding to inquiries, GPT may assist the chatbot in performing practical tasks immediately upon user request.

\subsection{Integration of Hybrid Model Components}
The proposed hybrid system follows a structured workflow in which each component has a specialized role: BERT for initial feature extraction, XGBoost for intent validation and refinement, optimized prompts for guiding GPT in DST tasks, and online agents for generating responses based on real-time information retrieval. By combining these elements, a unified architecture is created that makes use of each one's advantages. Key issues in conventional conversation systems, like preserving coherence, controlling ambiguity, and adapting to changing user input, are effectively handled by this approach by combining contextual understanding, intent validation, adaptive prompt-based generation, and dynamic information retrieval.

The proposed model combines BERT, XGBoost, GPT, and online agents to enhance DST by improving efficiency, precision, and user satisfaction. BERT provides contextual understanding, XGBoost validates and refines user intents, GPT generates responses, and online agents retrieve real-time information. Together, these elements create a cohesive system that effectively manages evolving user conversations while ensuring coherence and a natural user experience.

\subsection{Challenges Addressed in the Proposed Architecture}

By reviewing DST Challenges studied by \citet{Williams2016} and real-life dialogue challenges, several recurring issues in DST were identified, and targeted solutions were proposed for handling them which will be discussed in this section.

\subsubsection{Foundational Design and Language-Specific Adaptation}
To develop an efficient dialogue state tracker, it is vital to define the core concepts of the system and design a structured, well-defined baseline aligned with the model's objectives. While foundational dialogue concepts such as intents, slots, and states are general, their expression and interpretation can vary across each language due to differences in grammar, vocabulary, and linguistic complexity. Thus,  a customized baseline must be designed to handle each language specifically. 
Persian, a language spoken by nearly 130 million people worldwide, is the target of this research. To address this challenge, a comprehensive open-domain Persian dialogue dataset was created, key DST principles were modified to cope with the complex structure of Persian, and a system was developed to proficiently handle and monitor authentic Persian conversations.

\subsubsection{Input Scheduling for Effective DST}
One of the most fundamental challenges in conversational AI systems is determining how to format and deliver the input data to the trained model to maximize performance. This process is called a schedule in DST, the strategy for selecting and structuring the dialogue content sent to the model at each turn. After reviewing several methods, such as sending the whole conversation (both user and chatbot), only chatbot answers, etc., we realized the best possible schedule happens when we send user requests along with the intent of each turn, combined in the whole conversation. With this approach, we get the best results from the NLU, which was enabled to recognize almost all slot values and define the dialogue state.

\subsubsection{Handling Intent Shifts and Dynamic Context Transitions}
As conversational AI becomes more human-like, its complexity grows. One key challenge that emerges is the change of user's intents from one topic to another in the dialogue, which can cause the slot values to also change accordingly. For instance, two people can discuss the weather conditions of their city, Tehran, but end up talking about traveling to another city, Montreal. This can be a real challenge to the dialogue state tracker, as it must remain context-aware so it can define the state of the dialogue correctly. To address this challenge, a customized database was designed that saves the intent, slot value, and turn number of a dialogue. At each new turn, the input is passed to the NLU module both as a single input and in the context of the full conversation history. If the detected intent of that turn was different in comparison with the whole dialogue and database information, the tracker can admit a change in the intention of the dialogue and can act as the new intent of the user gets a response.

\subsubsection{Handling Unclear or Incomplete User Inputs}
In real-world scenarios, the user does not always provide a clear and complete answer. For instance, a user may ask for a recipe, but when the chatbot requests the specific dish, the user might reply with an unrelated or vague message or even nothing at all. In this case, even if the NLU can understand the potential intent shift, the proposed model must ask appropriate questions from the user to clarify. To address this challenge, the intent validator reviews the NLU's output and defines this situation as unclear. Due to this state, the model asks the user to clarify and prevents the chatbot to process the user's previous input unless it gets a clear answer. This policy guarantees the proposed model's accuracy for future responses.

\subsubsection{Disambiguation Strategies for Overlapping Intents}
Another key challenge that a dialogue state tracker must handle is making decisions through a doubtful scenario. This happens when the user's inputs reflect more than one intention. The user's primary goal is clear to them, but it may not be clear to the chatbot. Consider the example: "I wish there wasn't much distance between Toronto and Montreal; I'm so hungry, and I need to find a place for dinner." In this case, the user mentions both a geographical concern and a dining request within a single turn, leaving the chatbot in ambiguity as to which one must be responded to.  To handle this, the intent validator is trained on labeled data to detect ambiguous states. When an input is classified as ambiguous, the tracking process will be paused, and the chatbot will ask the user to clarify their main intent before proceeding.

\subsubsection{Handling don't care Scenario}
In some real-world interactions, users may not have a specific value for a slot in their mind. For example, a user might ask about cinema times, and when the chatbot requests the name of a specific movie, they simply respond with "whatever." This case is different from when the user responds unclearly or off-topic; the user is clearly saying that any value is acceptable, and they want the system to decide. In these cases, the model must recognize its status and choose a default (or random) value itself. Otherwise, the chatbot asks the user to clarify, creating confusion and leading to user dissatisfaction. The proposed system treats \texttt{don't care} as a valid and explicit intent. The NLU checks for this intent in each user turn, and if detected, a flag is raised, and the dialogue state tracker will choose a reasonable default value for that particular slot.

\section{Evaluation}
\label{sec:evaluation}

This section explores the experiments conducted to assess the efficacy of the proposed hybrid model for DST.

\subsection{Evaluation Metrics}Each module of this study was evaluated utilizing a variety of metrics. This section concentrates on the NLU component, intent validation, and the DST module. For the NLU module, slot accuracy was used to figure out the proportion of accurately predicted slot values, and intent detection accuracy was used to evaluate the model's capability in properly classifying user intentions. The intent validation module was examined for accuracy to see whether the validation process successfully confirmed the recognized intent. 

For DST, Joint Goal Accuracy (JGA), Flexible Goal Accuracy (FGA), and Average Goal Accuracy (AGA) were employed which are defined as follows:

\begin{itemize}
\item \textbf{JGA}: Measures the accuracy of accurately predicting all parameter values throughout a certain dialogue.

\item  \textbf{FGA}: Assesses the model's capacity to properly predict the dialogue state despite possible mistakes in some parameters during the conversation.

\item  \textbf{AGA}: Evaluates the mean accuracy of various parameters during the dialogue, providing a broader view of the model's ability to observe each piece of information.
\end{itemize}

\subsection{Setup of Experiments}The goal of the experimental design was to evaluate the accuracy of the model in both practical and monitored scenarios utilizing GPU-accelerated computation. The key components of the experimental setup are as follows:

\begin{itemize}
\item  \textbf{NLU}: RoBERTa-large was fine-tuned on a Persian multi-turn dialogue dataset based on the MultiWOZ structure to implement language-specific characteristics accurately.

\item \textbf{Intent} \textbf{Validation}: XGBoost was trained on features derived from NLU's outputs to validate and strengthen intent detection, guaranteeing precise and dependable detection of user intents.

\item \textbf{GPT Configuration for Dialogue } \textbf{DST}: GPT was instructed with prompt engineering tailored for diverse intentions to execute the DST task. Engineered to ensure that GPT generates optimized outputs, these prompts include vital parts consisting of dialogue states, SQL queries, and follow-up questions for the chatbot to lead the conversation.

\item  \textbf{GPT for Online Agents}: GPT served as the core model for developing online agents that dealt with real-time customer inquiries. These agents dynamically sourced information from external entities and generated responses that incorporated this data into coherent and contextually pertinent answers. This functionality facilitated the system for the evolution of user-specific contexts and knowledge bases.

\item \textbf{Evaluation Protocol}: Cross-validation was employed to ensure robustness over several test partitions. To evaluate each of the system's components individually, a variety of metrics were implemented:
\begin{itemize}
\item In the NLU module, slot accuracy and intent classification accuracy were examined to evaluate the model's competence in effectively recognizing slot values and classifying user intentions.

\item  In the intent validation module, accuracy was utilized to evaluate the confidence of validating expected intents.

\item  JGA, FGA, and AGA scored the DST model's performance in keeping track of the dialogue state.
\end{itemize}

\item  \textbf{Infrastructure}: To facilitate the efficient execution of large-scale computations, the entire study was conducted on a high-performance system equipped with an NVIDIA RTX 3090 GPU (24 GB of VRAM).

\item \textbf{Real-Time Components}: Online agents were able to generate accurate and up-to-date responses that were personalized to user requests by collaborating effectively with other modules, ensuring the real-time retrieval and integration of external information.
\end{itemize}

\subsection{Results}
The results are presented in this section, showing that the proposed model has reached excellent performance in NLU, intention validation, and DST, which are the three main areas evaluated. Each module was assessed using a wide range of metrics.

\subsubsection{Natural Language Understanding}

The NLU component for slot filling and intent detection was implemented leveraging a fine-tuned RoBERTa model. This module was evaluated on our Persian multi-turn dialog dataset, achieving remarkable performance across all key metrics.

\begin{table*}[!h]
\caption{{NLU Performance (10-fold Average)} }
\label{table-wrap-d85b531bfccc4f2f84973ea10c087ec3}
\def\arraystretch{1}
\ignorespaces 
\centering 
\begin{tabular}{l l l l}
\hline
\textbf{Task} &
  \textbf{Acc.} &
  \textbf{F1} \textbf{micro} &
  \textbf{F1} \textbf{macro}\\ \hline \hline
 Intent Detection &
   96.13 &
   96.13 &
   92.68\\
 Slot Filling &
   99.56 &
   97.35 &
   91.35\\
   \hline
\end{tabular} 
\end{table*}
As shown in Table 1, the intent detection ability was completely reliable since the model recognized approximately all the user intentions across many dialogues, with an accuracy of \textbf{96.13\%}. Slot filling also performed perfectly, reaching a token-level accuracy of \textbf{99.56\%}. However, a slight gap between F1 Micro and F1 Macro scores in slot filling (97.35\% vs. 91.35\%) indicates marginally lower performance on infrequent slot types, an expected outcome in imbalanced datasets. Overall, the NLU module shows significant robustness in both tasks.

\subsubsection{Intent Validation}

To validate the NLU module's intent predictions, a classification task was applied to 49,694 outputs. These were manually categorized into three classes: \texttt{confirmed}, \texttt{unclear}, and \texttt{ambiguous}. Since 96.97\% of the data was labeled as confirmed, this led to a significant class imbalance challenge. Furthermore, some prediction scores, such as smoothed max scores from the NLU module, could not be directly used for pattern recognition.

For handling these challenges, two machine learning algorithms were evaluated: \textbf{random forest }and \textbf{XGboost}. Random Forest was first trained using grid search to optimize parameters after the data had been standardized. Its performance is shown in Table 2.

\begin{table*}[!htbp]
\caption{{Random Forest Results} }
\label{table-wrap-63e542e49c9a40fd80231c576116bbbe}
\def\arraystretch{1}
\ignorespaces 
\centering 
\begin{tabular}{l l l l}
\hline
\textbf{Label} &
  \textbf{Precision} &
  \textbf{Recall} &
  \textbf{F1 Score}\\ \hline \hline
 Confirmed &
   0.97 &
   1.00 &
   0.99\\
 Ambiguous &
   0.43 &
   0.03 &
   0.06\\
 Unclear &
   0.29 &
   0.05 &
   0.09\\
   \hline
\end{tabular}
\end{table*}
Even though the model's performance on the majority class (\texttt{confirmed}) was remarkable, its performance on minority classes (\texttt{ambiguous}, \texttt{unclear}) was far from ideal. To improve classification on minority labels, an \textbf{XGboost }model was trained, which is known for handling imbalanced data more effectively. This model's results are shown in Table 3.

\begin{table*}[!htbp]
\caption{{XGBoost Results} }
\label{table-wrap-d2f437a8ef57469c894856e01507b0cc}
\def\arraystretch{1}
\ignorespaces 
\centering 
\begin{tabular}{l l l l}
\hline
\textbf{Label} &
  \textbf{Precision} &
  \textbf{Recall} &
  \textbf{F1 Score}\\ \hline \hline
 Confirmed &
   0.97 &
   1.00 &
   0.99\\
 Ambiguous &
   0.46 &
   0.14 &
   0.21\\
 Unclear &
   0.43 &
   0.10 &
   0.17\\ \hline
\end{tabular} 
\end{table*}
This model significantly outperformed the random forest algorithm on minority labels. However, it still presents a strong bias toward the dominant class. To handle this challenge, feature engineering was performed over the NLU prediction scores (e.g., smoothed max and normalized values), polynomial feature interactions, class weighting for minority sensitivity, and threshold tuning to balance class detection to enhance the XGboost model. This optimized model achieved more balanced predictions, as shown in Table 4.

\begin{table*}[!htbp]
\caption{{Improved XGBoost Results} }
\label{table-wrap-0a1e9ef910a94a8cae00f6cbb4f1788d}
\def\arraystretch{1}
\ignorespaces 
\centering 
\begin{tabular}{l l l l}
\hline
\textbf{Label} &
  \textbf{Precision} &
  \textbf{Recall} &
  \textbf{F1 Score}\\ \hline \hline
 Confirmed &
   0.98 &
   0.96 &
   0.97\\
 Ambiguous &
   0.19 &
   0.34 &
   0.24\\
 Unclear &
   0.21 &
   0.33 &
   0.26\\
   \hline
\end{tabular}
\end{table*}
As shown, while precision and recall for the \texttt{confirmed} class dropped slightly, a considerable boost in performance can be seen for both the \texttt{ambiguous} and \texttt{unclear} categories. This reflects a more balanced model capable of identifying all intents, better aligning with the chatbot's validation requirements.

\subsubsection{Dialogue State Tracking}The DST module's evaluation was done with a set of well-known dialogue-based evaluation metrics in conversational AI research. To assess the model with the main metrics these benchmarks were considered:

\begin{itemize}
\item  \textbf{Correct Intent Detection}: Ability to accurately detecting the user's intent.

\item  \textbf{Correct Slot Filling}: Correctly extracting slot values from user inputs.

\item \textbf{Whatever State Detection}: Recognizing when the input does not indicate any particular value.

\item \textbf{Correct}\textbf{\space }\textbf{State} \textbf{Recognition}: Accurately recognizing every dialogue state through each section of the conversation.

\item  \textbf{Human-Evaluated}\textbf{\space }\textbf{Quality}\textbf{\space }\textbf{of} \textbf{DST}: Overall quality as judged through manual annotation.

\end{itemize}

One of the most widely used metrics in evaluating dialogue systems is the \textbf{JGA}. This metric reflects how well the system identifies all benchmarks throughout each turn. This rating is a strict measure since it will fail entirely whenever any of the benchmarks are predicted incorrectly. Therefore, JGA is often seen as a comprehensive benchmark of a DST system's reliability \cite{Dey2022}. Using this metric, the proposed DST model achieved a JGA of \textbf{0.73}. This score reflects robust performance, especially under such a strict metric.

Unlike JGA, which requires exact matches across all benchmarks, FGA allows for partial correctness. It considers a prediction successful if at least one of the benchmarks is correctly fulfilled.

This flexibility is particularly useful in real-world scenarios where dialogues may include incomplete or noisy information. FGA, therefore, acknowledges partially correct responses as successful, helping evaluate models more pragmatically while still maintaining relevance to the dialogue objective \cite{Dey2022}. Using this definition, the model achieved a perfect FGA score of \textbf{1.0}, meaning that for every input, the system produced at least one correct prediction.

The AGA metric independently assesses the model's ability to satisfy each of the five evaluation benchmarks and then computes the mean accuracy across them. AGA accuracy gives us a general understanding of the model's performance since it considers all benchmarks separately, despite the previous metric. As a result, AGA is very effective for evaluating complex multi-slot intents. Using this metric, the proposed DST model achieved an AGA of \textbf{0.92}. This high score suggests strong general reliability across all sub-tasks, confirming the model's robustness \cite{Dey2022}.

Table 5 presents the overall results of the DST module.

\begin{table}[H]
\centering
\caption{Evaluation Results for the Proposed DST Model}
\label{tab:dst-metrics}
\begin{tabular}{l c}
\toprule
\textbf{Metric} & \textbf{Score} \\
\midrule
Joint Goal Accuracy (JGA) & \textbf{0.73} \\
Flexible Goal Accuracy (FGA) & \textbf{1.00} \\
Average Goal Accuracy (AGA) & \textbf{0.92} \\
\bottomrule
\end{tabular}
\end{table}

\section{Conclusion}
\label{sec:conclusion}

This paper presented a hybrid architecture model for Persian-based chatbots, including NLU, intent validation, and DST modules. By tackling common problems like unbalanced data, complex multi-slot intent structures, and slot dependencies, the proposed model showed a strong ability to comprehend and handle conversations.

Future endeavors will concentrate on expanding the dataset with a greater variety of domain-rich Persian dialogues, enhancing model flexibility via multilingual training, and including user-centric features such as emotion recognition. For long-term success in practical implementations, it will also be crucial to integrate continuous learning based on user input and extend evaluation to more open-ended conversational areas.
    
\section*{Declaration of generative AI and AI-assisted technologies in the writing process}

During the preparation of this work, the author(s) used ChatGPT, QuillBot, and Grammarly in order to enhance language clarity, rephrase sentences for improved readability, and correct grammatical errors. After using these tools, the author(s) reviewed and edited the content as needed and take(s) full responsibility for the content of the publication.


 \bibliographystyle{elsarticle-num-names}
 \bibliography{references}



\end{document}